\documentclass[twoside,twocolumn]{article}

\usepackage{color}
\usepackage{amsmath}
\usepackage{amsfonts}
\usepackage{amssymb}
\usepackage{cite}
\usepackage[dvips]{graphicx}
\usepackage{amsthm}
\usepackage{caption}

\usepackage{algorithm}
\usepackage{algorithmic}

 \newtheorem{remark}{Remark}

\begin{document} 

\title{Bending the Curve: Improving the ROC Curve Through Error Redistribution}

\author{
Oran Richman\\ 
Department of Electrical Engineering
Technion
Haifa, Israel\\ 
\texttt{roran@tx.technion.ac.il} \\
Shie Mannor \\
Department of Electrical Engineering
Technion
Haifa, Israel \\
\texttt{shie@ee.technion.ac.il} 
}









\maketitle
\begin{abstract}
 Classification performance is often not uniform over the data. Some areas in the input space are easier to classify than others. 
Features that hold information about the "difficulty" of  the data may be non-discriminative and are therefore disregarded in the classification process. We propose a meta-learning approach where performance may be improved by post-processing. This improvement is done by establishing a dynamic threshold on the base-classifier results. Since the base-classifier is treated as a ``black box'' the method presented can be used on any state of the art classifier in order to try an improve its performance. We focus our attention on how to better control the true-positive/false-positive trade-off known as the ROC curve. We propose an algorithm for the derivation of optimal thresholds by redistributing the error depending on features that hold information about difficulty. We demonstrate the resulting benefit on both synthetic and real-life data.
\end{abstract}

\section{Introduction} \label{section:Introduction}

Binary classification is perhaps the most widely studied in machine learning and many methods are used to obtain binary classifiers from data. For most applications two performance measures are of special interest. The first is the True Positive Rate (TPR)--the portion of true positives that are classified as such by the classifier. The second is the False Positive Rate (FPR)--the portion of true negatives that are classified as positive by the classifier.

 There is a fundamental trade-off between those two measures. This trade-off is often controlled through thresholding: the classifier produces a continuous score for each sample,  and a threshold is used to determine if the sample is classified as positive (above the threshold) or negative (below the threshold).  The pair (FPR,TPR) is the {\it operating point} of the resulting classifier.

  The typical approach is to vary the threshold and obtain the complete curve of operating points called the Receiver operating characteristic (ROC) curve \cite{hanley1982meaning}.
The performance of the classifier is then evaluated based on the whole curve using a specific operating point (i.e., a desired FPR level) or by considering the area under the curve (AUC).
The area under the curve is an interesting measure since it as a probabilistic interpretation. The area under the curve of a classifier $h(x), \mathbb{R}^n \to \mathbb{R} $ is the probability that for a random positive sample $x^+$ and a  random negative sample $x^-$ the classifier will produce $h(x^+)>h(x^-)$.
In this paper we show that the thresholding approach can be refined such that performance can be improved  {\em without retraining the classifier}. 


 Our approach is based on two observations. The first is that even after conditioning on the true class of the sample, the score is often correlated with some features (we will refer to them as auxiliary features).  Moreover, those features may hold no or little  discriminative information and  are therefore disregarded during the learning process.  For example, picture resolution may affect performance of  object recognition greatly \cite{hoiem2012diagnosing}. It is however often uncorrelated to the picture content. The Discriminatingly Trained Deformable Part Model classifier \cite{voc-release5} is a popular state of the art object detector. It can be seen that in this classifier high resolution pictures receive higher scores compared with low resolution pictures \cite{pedersoli2014toward}. 
 
 The second observation is that the correlation with the score of positive examples and the correlation with the score of negative examples  may be  statistically different and even significantly so. We are mainly concerned with features that are correlated with the  ``difficulty'' of the problem. The reference to ``difficulty'' implies some different effect of those features on the positive and negative examples score.  For example, the scores of the positive and negative examples get more or less concentrated. Revisiting the image resolution example, the effect or reducing resolution on a real-object's score differs from the effect on a random background image. This difference can be exploited to improve performance for a specific operating point.
  
For every desired operating point, we propose to use a threshold that depends on auxiliary features instead of being fixed for the entire input-space: the threshold is a {\em function} instead of a constant as in the standard approach. The threshold ``curve'' can be designed so that performance is improved (i.e., higher TPR for a given FPR or a lower FPR for a given TPR). Our approach effectively rebalances the performance in different areas of the input space and redistributes the error.
 
A simple heuristic for  determining the threshold (as a function of the features) is to eliminate the correlation between the adjusted score (original score difference from the threshold) and the features. However, in the case where the positive and negative samples are affected differentially this is not trivial and requires estimating the conditional distribution of each class given the features.

\begin{figure}
      \centering
      \centering
      \captionsetup{justification=centering,margin=1cm}
      \includegraphics[angle=0,
           width=\linewidth,height=3in]{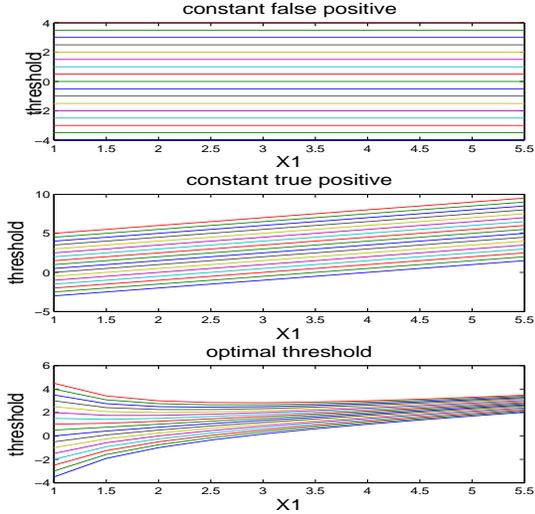}
            \caption{threshold as a function of $X_1$ for different approaches and for different operating points (Example 1) }
            \label{fig:threshold}
            \end{figure}

 The score can be adjusted either according to the positive examples or according to the negative examples.
In the first case we   use a threshold which follows the mean of the score of negative examples. We refer to this approach as ``constant false positive rate". 
Another approach is to use a threshold that follows the mean of the score of positive examples.  We refer to this approach as ``constant true positive rate". 
An illustration of these approaches on a simple example can be seen on Figure \ref{fig:threshold}. 

Both approaches, however, suffer from the same structural deficiency, some threshold ``curve'' is derived and then the entire ROC curve is created by adding a fixed offset to it.

 We present the Optimal Error Redistribution (OER) framework that ``bends'' the curve differently for different operating points. Our method is general and does not require any knowledge concerning the learning process used to train the classifier. The classifier is treated as a ``black box'' allowing to 
``bend the curve" for a wide variety of classifiers.
  
 Our method is based on an alternative view of the ROC curve. Instead of viewing the operating point as a consequence of a varying threshold, we can consider the following optimization: Given some desired FPR, find the threshold curve (threshold as a function of the auxiliary features) that brings the TPR to a maximum.  This essentially treats the FPR as a resource which need to be distributed between samples. Easy examples will contribute (in expectation) lower FPR than that contributed by the harder examples. This view allows introducing methods from the field or resource allocation (for example, methods from sensor management; a good review can be found at \cite{hero2011sensor})

 \paragraph{Example 1}
  
   Consider the following case. Some random variable $X_1$ is drawn uniformly from the set $[1 \;, 5]$. Some random variable $Y \in \{-1,1\}$ is drawn such that $Y=1$ with probability 0.5. The random variable  $X_2$ is then drawn according to the following distribution:
  \begin{equation*}
  X_2|y=1 \sim N(X_1,1),\;\;\;  X_2|y=-1 \sim N(0,1) 
  \end{equation*}
  
  Since $X_1$ contains no discriminative information, a reasonable classifier for Y is $h(X_1,X_2)=X_2$ (using a linear classifier do not change the results significantly, however it makes the visual understanding of the following figures more difficult).  Figure \ref{fig:threshold} shows dynamic thresholds (with respect to $X_1$) derived from the different approaches described above. The upper figure shows the curve matching the constant false positive approach. In this example it coincides with the original fixed (with respect to $X_1$) threshold. The middle figure shows the curve matching the constant true positive approach. This corresponds to a linear classifier which uses also the data in $X_1$. Both threshold curves are not optimal. The lower figure show the optimal curves. It can be seen that for different operating points the curve ``bends''. When the example is ``hard'' to classify, the optimal threshold varies much more than when the example is ``easy''. Using a more complex classifier may produce different curves than those presented in those figures but will not be able to produce the ``bending'' effect. 
  
   \begin{figure}
      \centering
      \includegraphics[angle=0,
      width=\linewidth,height=3in]{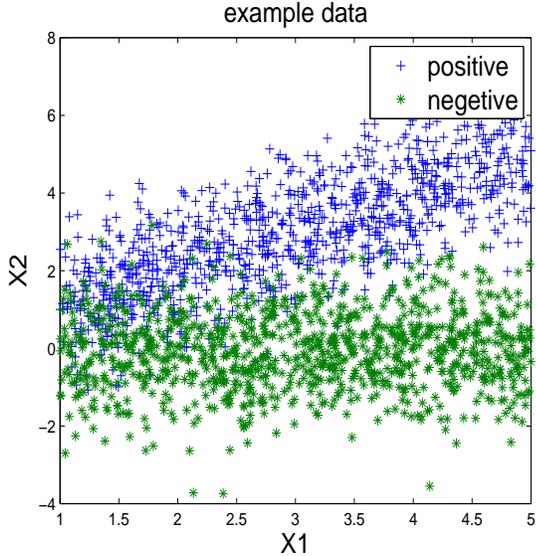}
      \caption{Data distribution of Example 1}
      \label{fig:data}
      \end{figure}
    
  \paragraph{Example 2}
  
\begin{figure}
                
               \centering
               \captionsetup{justification=centering,margin=1cm}
                   \includegraphics[angle=0,
                   width=\linewidth,height=3.1in]{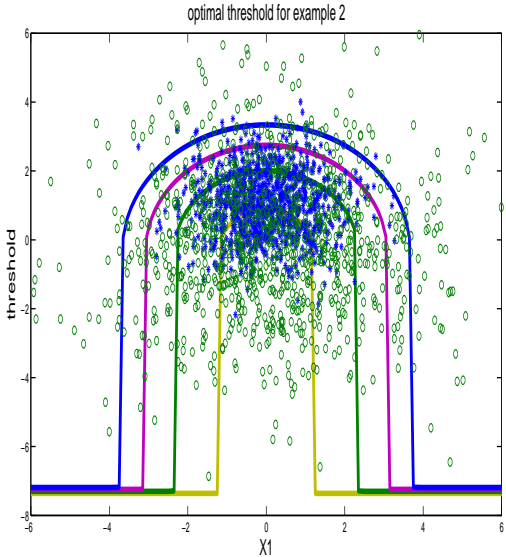}
                   \caption{Data distribution and some optimal thresholds for Example 2}
                   \label{fig:threshold-ex2}
                
          \end{figure}

  The optimal threshold may vary even when the mean and standard-deviations do not depend on the features. This may happen when the prior changes. Meaning, the ratio between the quantity of positive and negative examples is related to the auxiliary features.  As an  example, consider the following. Some random variable $Y \in \{-1,1\}$ is drawn such that $Y=1$ with probability 0.5. Some random vector $x=(X_1, X_2)$ is then drawn according to the following distribution: 
  \begin{equation*}
  x|y=1 \sim N((0,1),2I),\;\;\;  x|y=-1 \sim N((0,0),I),
  \end{equation*}
  where $I$ is the 2x2 unit matrix. It is easy to see that a reasonable classifier for Y is $h(X_1,X_2)=X_2$
    Observe  that in this example the constant true positive and constant false positive coincide and derive a constant threshold. 
  Figure \ref{fig:threshold-ex2} shows the data distribution of this example along with some optimal dynamic thresholds (with respect to $X_1$). It can be seen that our method had essentially created a non-linear classifier for each desired operating-point. As before, the different curves are not with fixed offset from one another.  Interestingly, for large enough $|X_1|$ the prior is so significant that the optimal threshold is at $-\infty$. This characterizes situations in which the standard deviation of the positive examples score is larger than that of the negative examples score.  
  
\paragraph{Related work}

Meta learning   \cite{vilalta2002perspective} is concerned with the enhancement of classifiers. A meta classifier takes a set of classifiers (base classifiers) and merges them in various ways to produce a unified classification result.  The base classifiers are often trained using some variations of the same training set. This includes bagging \cite{breiman1996bagging}, boosting \cite{freund1996experiments} and many others (for example \cite{breiman1996stacked,ho1998random}). 
 Some works in this field target specifically the improvement of the ROC curve. In \cite{provost2001robust} the authors proposed the ROC Convex Hull (ROCCH) method. The ROCCH is based on the observation that given two classifiers with different ROC curves any point on the line segment between two operation points can be achieved. This is by randomly using one or the other classifier with appropriate probabilities. This allows to combine several classifiers to achieve an ROC curve which is better than each classifier. The method we are presenting in this paper shares the basic approach with the field of  meta-learning. In our case, the set of base-classifiers is the base classifier with different thresholds. 
 
 We differ, however, from existing work in this field  in two important aspects. First, we use as input  only a single classifier. Second, the auxiliary feature space can be completely different from the base classifier's feature space. We do not require any  ``re-training'' and no access to the classifier inner-workings is needed. As a result our method is much less sensitive to the way the original classifier was derived. This  allows, in our view, much greater flexibility in applying this method. Note that we do require some training set to determine the dynamic thresholds. This set however can be different from the one used to train the classifier. 

A different approach that targets specifically the improvement of the ROC curve is trying to build classifiers that optimize the area under the curve (AUC) directly \cite{cortes2004auc,yan2003optimizing,herschtal2004optimising}.  Using various surrogates the area under the curve can be optimized to derive some $h^{opt}(x)$. The optimization is done with respect to some hypothesis class. Our method does not optimize the AUC but rather optimizes the ROC curve point by point. The resulting classifier however is in a different hypothesis class than the base classifier. The relation between  $h^{opt}(x)$ and the result of using our method on some $h(x)$ is unclear. This is since the optimization of both methods is done for different hypothesis classes. However, our method can even improve $h^{opt}$ after it is derived using one of the AUC optimization methods.  It is important to note also that while optimizing the AUC is possible for some limited set of hypothesis classes our method is general and can accommodate complex learning schemes.

Recent work has also explored different threshold choice methods \cite{hernandez2012unified,drummond2006cost,fawcett1997adaptive}. A threshold choice method adjusts the threshold to accommodate changes in the cost functions or class distributions. Those methods share a similarity with the ideas presented in this paper. However, the setting which we explore in this paper is substantially different. In our setting the threshold may vary between different regions of the input-space with the goal of achieving maximal \emph{average} performance. The above mentioned work explores the case where the threshold is used to adapt the base classifier in order to maximize \emph{current} performance.  

It is important to note that simply appending the auxiliary features to the features vector will not produce the same result. First, similarly to meta learning, the resulting hypothesis class is significantly larger than that of the base classifier. Moreover, in many cases it is far from trivial to parametrize the resulting hypothesis class in such a way that will allow learning a ``standard'' classifier. As can be seen by examples 1 and 2, using the method presented allows creating complex classifiers using simple (linear) base classifier. This also implies that simply treating the auxiliary features as features will often provide much smaller benefit. Also, it is far from trivial to directly learn such complex hypothesis classes.   


It is possible, obviously, to incorporate the ideas presented in this paper in the learning process of the base classifier. While such tight-coupling  may produce better results such adoption is far from being trivial for most learning schemes. 
The method presented is treating the classifier as a ''black box". Therefore, it can be easily incorporated on top of any existing classifier. As mentioned before, in our method the threshold is a function of the auxiliary features. If the base-classifier was ``smart enough'' to use the full information contained in those features then the method will produce no benefit. As we will see in the following this is often not the case, especially when the features have low correlation with the real class.

Our contributions are threefold: First, we introduce a novel framework in which the threshold may vary over the input-space. Second,  we introduce the Optimal Error Redistribution (OER) method. This method allows the creation of a meta classifier with improved ROC curve comparing with the base classifier. In addition,  we derive a closed form solution of the optimal threshold for the special case of Gaussian distributions. Simulations which demonstrate the benefit which may arise are presented.  Finally, we present a feature selection technique (for OER). This allows the selection of the auxiliary features without the explicit calculation of the ROC curve.

We believe that the method presented in this paper should become a standard tool in ROC analysis. It is always beneficial to try and improve the ROC curve some more and our method proposes a generic way to do so.

This paper is structured as follows: Section \ref{Model Formulation} defines the problem formally and provides the general OER method. Section \ref{sec:gauess} details a simple implementation and provides a closed-form solution for a special case. Section \ref{sec:features} outlines a feature selection technique which allows to select features for the method without the explicit calculation of the ROC curve.  Section \ref{Simulation Results} demonstrates the feasibility of the problem on real-life data while Section \ref{section:conclusion} concludes the paper with some final thoughts and some still open questions.   

\section{Optimal Error Redistribution} \label{Model Formulation}

Consider  binary classification of objects represented by some vector $x\in \mathbb{R}^n$.
The  base classifier is based on some function $h(x), \mathbb{R}^n \to \mathbb{R} $. In the original classification scheme a threshold is used to transform the output of the function to a binary classification. A sample is classified as positive if $h(x)\geq k$ and negative otherwise.  We allow the threshold to depend on some auxiliary feature vector $\tilde{x}$. Notice that  $\tilde{x}$ should not be confused with the vector $x$ that represents the data. The feature vector $\tilde{x}$ can be some subset of $x$ or measured separately from the raw data (as in the example of picture resolution).

 We would like to find some function $k(\tilde{x})$ which assigns a threshold for each example.  We approximate this function by partitioning the feature space into $N$ bins. Each bin can be assigned a different threshold. The determination of a continuous function $k(\tilde{x})$ is possible in a special case which is outlined  in Section \ref{Constant Variance}.   
Formally, The data distribution is modelled as a superposition of $N$ populations $\{A_i \; i=1,\ldots,N\}$. The auxiliary feature vector $\tilde{x}$ deterministically determine the population from which the example was taken. 
In the derived meta-classifier the original scalar threshold $k$ is replaced with a vector $(k_1,\ldots,k_N)$. Sample $x\in A_i$ is classified as positive if  $h(x)\geq k_i$ and negative otherwise. 

In each population the score distribution obeys the following:
\begin{equation}\label{data_distribution}
\begin{array}{ccc}
h(x)|x\in A_i, y=1& \sim& f_i,\\
h(x)|x\in A_i, y=-1 &\sim& g_i.
\end{array}
\end{equation}
Where $f_i$ and $g_i$ are probability density functions. Denote the corresponding cumulative distribution functions as $F_i$ and $G_i$.
Further, $p_i^+=\mathbb{P}(x\in A_i|y=1)$ and $p_i^-=\mathbb{P}(x\in A_i|y=-1)$.

The optimal threshold curve is given by solving an optimization problem in which the average TPR is maximised while some constraint is imposed on the average FPR. Namely, the optimization problem: 

\begin{equation}\label{gen_prob}
\begin{array}{l}
\max_{(k_1,\ldots,k_N)} \sum\limits_{i=1}^{N} p_i^+(1-F_i(k_i))\\
s.t\\
	\sum\limits_{i=1}^{N} p_i^-(1-G_i(k_i))=C.
\end{array}
\end{equation}  

This problem can be non-concave and finding the global maximum may be hard \cite{rao2009engineering}. We can however, use an equivalent form of problem (\ref{gen_prob}) to construct a gradient ascent algorithm that will lead us to a local maximum. Instead of solving Problem (\ref{gen_prob}) we will solve the following problem for some $\lambda>0$: 

\begin{equation}\label{gen_prob_eq}
\begin{array}{l}
\max_{(k_1,\ldots,k_N)}  \sum\limits_{i=1}^{N} p_i^+(1-F_i(k_i))-\lambda\sum\limits_{i=1}^{N} p_i^-(1-G_i(k_i)).\\
\end{array}
\end{equation}  

It is known that for both problems a necessary condition for a vector $(k_1,\ldots,k_N)$ to be a solution is given by 
\begin{equation} \label{gen_problem_necessary_condition}
p_i^+f_i(k_i)=\lambda p_i^-g_i(k_i).
\end{equation}

We will denote as the benefit-cost ratio the expression:
\begin{equation}\label{benefit-cost}
\frac{p_i^+f_i(k_i)}{p_i^-g_i(k_i)}.
\end{equation} 
For a thresholds vector  $(k_1,\ldots,k_N)$  to be optimal the benefit-cost ratio should be constant between populations.
 
The OER algorithm  is given by Algorithm \ref{alg:OER}.   As we will see in the following, for the special case where $f_i$ and $g_i$ are Gaussian with the same variance, it is possible to derive a closed-form solution for the global maxima.   The necessary condition (\ref{gen_problem_necessary_condition}) implies that for the optimal threshold the benefit-cost ratio is constant between populations.
 
Notice that since we would like to derive the complete ROC curve there is no need to solve the problem for different values of $C$. We can use the common benefit-cost ratio $\lambda$ as a parameter and derive the ROC curve by varying $\lambda$. A specific operating point can then be chosen for implementation. 

\begin{algorithm}[tb]
   \caption{OER}
   \label{alg:OER}
\begin{algorithmic}
   \STATE {\bfseries Paramters}    $\zeta$ - learning rate, $\epsilon$ - stopping threshold.
   \STATE {\bfseries Input:}    $f$,  $g$, $p^+$ ,$p^-$, $\lambda$  
   \STATE all vector operations are done point-wise.
   \STATE $\Delta=1$
   \STATE $k=(0,0,\ldots,0)$
      
   \WHILE {$\Delta>\epsilon$}
   \STATE  $k=k-\zeta[p^+f(k)-\lambda p^- g(k)]$
   \STATE $\Delta=||p^+f(k)-\lambda p^- g(k)||_2$
      
   \ENDWHILE
   \STATE return k
   \end{algorithmic}
\end{algorithm}

The method presented can be also interpreted from a calibration perspective. Calibration is used to transform classifier outputs into posterior probabilities \cite{platt1999probabilistic,hastie98classificationby}. One popular calibration method, known as Platt calibration, fits a sigmoid model to the data \cite{platt1999probabilistic}. The method finds two parameters $a$ and $b$ such that the posterior probability fits as good as possible to $P(y=1|h(x))=\frac{1}{1+exp(ah(x)+b)}$. Earlier work as used a Gaussian fit as the base distribution \cite{hastie98classificationby}. 

Our method (with a slight modification, since we also use Gaussian as our base distribution) can be viewed as an extension to Platt calibration where the two scalars $a$ an $b$ are replaced with two functions of the auxiliary features. This results in:
\begin{equation*}
P(y=1|h(x))=\frac{1}{1+exp(a(\tilde{x})h(x)+b(\tilde{x}))}.
\end{equation*}
The posterior probabilities can then be compared to a threshold such that the resulting classifier is equivalent to that received by our method. while this interpretation of our method is valid we believe that the interpretation detailed in this paper is clearer and easier to implement. Some previous work by Vapnik \cite{vapnik1998statistical}  considered a calibration method which is not uniform over the sample space. However, this method is limited to Support Vector Machines (SVM) and uses the original feature space with no auxiliary features. Our method is much more general.  


\section{Implementation} \label{sec:gauess}

The OER method presented earlier is general and flexible. There are two main design choices. First choosing the auxiliary features and corresponding $A_i$. Second, choosing a model for $f_i(y)$ and $g_i(y)$ and a corresponding method for fitting the data. Section \ref{sec:features} provides a heuristic method for choosing auxiliary features. However, this question is still open and a topic for future research. One simple model for $f_i(y)$ and $g_i(y)$ can be the use of a Gaussian model for the conditional behaviour of the score. 
Formally, the Gaussian model is stated as:
\begin{equation*}
\begin{array}{l}
  h(x)|x\in A_i, y=1 \sim N(\mu_i^+,\sigma_i^+) \\ 
  h(x)|x\in A_i, y=-1 \sim N(\mu_i^-,\sigma_i^-).
\end{array}
\end{equation*}
One of the main benefits of using such a model is that it requires only the estimation of the first and second moments. Both can be easily estimated for each bin. 
%
%




 The necessary condition for an extremum now takes the form of:
\begin{equation} \label{necessary-condition} 
\frac{p_i^+}{\sigma_i^+}e^{-\frac{{(k_i-\mu_i^+)}^2}{2{\sigma_i^+}^2}}=\frac{p_i^-}{\sigma_i^-}\lambda e^{-\frac{{(k_i-\mu_i^-)}^2}{2{\sigma_i^-}^2}}.
\end{equation}
Where $k_i$ is the threshold for the desired classifier.

The benefit-cost ratio is 
\begin{equation} \label{gaues-benefit-cost-ratio}
\frac{p_i^+\sigma_i^-}{p_i^-\sigma_i^+} e^{-\frac{({(k_i-\mu_i^+)}^2}{2{\sigma_i^+}^2}+\frac{({(k_i-\mu_i^-)}^2}{2{\sigma_i^-}^2}}.
\end{equation}

\begin{figure}
  \centering
  \includegraphics[angle=0,
  width=\linewidth,height=3in]{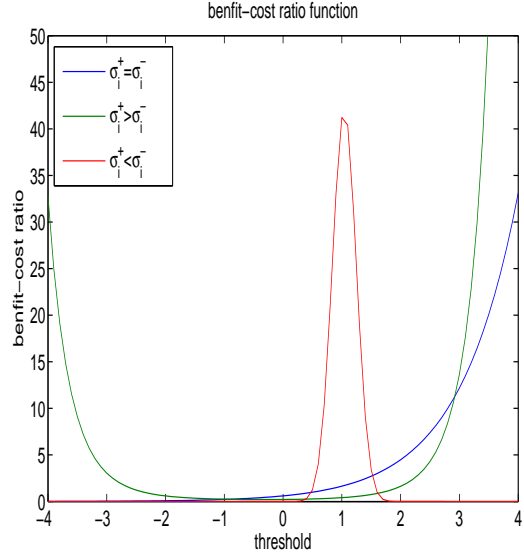}
  \caption{Benefit-cost ratio for different parameters of a Gaussian distribution}
  \label{fig:benfit-cost}
  \end{figure}

 An illustration of the benefit-cost ratio can be seen in Figure \ref{fig:benfit-cost} for different relations between  $\sigma_i^+$ and $\sigma_i^-$ .
 Notice that when $\sigma_i^+=\sigma_i^-$ the ratio (\ref{gaues-benefit-cost-ratio}) is strictly monotone in $k_i$. Therefore, if $\forall i \; \sigma_i^+=\sigma_i^-$ then (\ref{necessary-condition}) admits a single solution for every $\lambda$. In that case,  a closed form solution to the optimization problem can be derived. This however is not the case in general. In the general case multiple extremum points may exist and therefore  local optimization methods need to be used. Notice also that if $\sigma_i^+>\sigma_i^-$  then there is a minimum to the benefit-cost ratio. Therefore, for large enough FPR the optimal threshold is $-\infty$. Similarly if $\sigma_i^+<\sigma_i^-$ then there is a maximum to the benefit-cost ratio and for small enough FPR the optimal threshold is $\infty$. 

A solution for the optimization problem can be found by using OER (Algorithm \ref{alg:OER}). The gradient is given by: 
 \begin{equation}
 \nabla_i=\frac{p_i^+}{\sigma_i^+}\exp(-\frac{(k_i-\mu_i^+)^2}{2{\sigma_i^+}^2})-\frac{p_i^-\lambda}{\sigma_i^-}\exp(-\frac{(k_i-\mu_i^-)^2}{2{\sigma_i^-}^2}).
 \end{equation}  
 
 Notice that if $\sigma_i^+>\sigma_i^-$ then the optimal threshold may be $-\infty$ and if $\sigma_i^+<\sigma_i^-$ then the optimal threshold may be $\infty$. It is advised at each step to project the threshold into some fixed interval $[-K \; K]$  such that the gradient-ascent method will converge.
 
 \paragraph{Example 2 revisited}
 
\begin{figure}
       \centering
           \captionsetup{justification=centering,margin=1cm}
              \includegraphics[angle=0,
              width=\linewidth,height=3in]{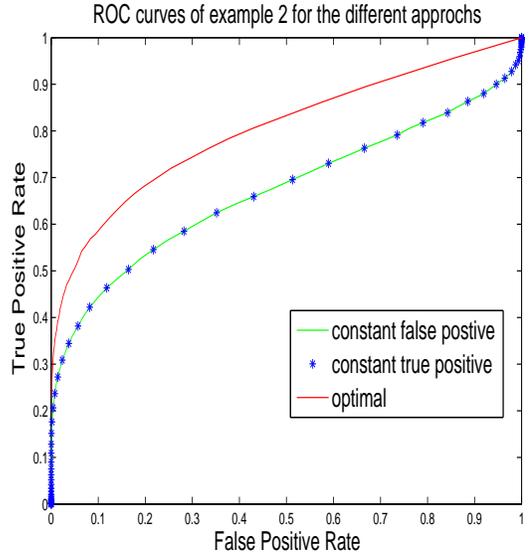}
              \caption{ROC curve of Example 2 }
              \label{fig:example2}
       \end{figure}

 We have used the described method on example 2. We divided $X_1$ values into 120 bins, where $x\in A_i$ if $-6+0.1i<X_1<-5.9+0.1i$. Two additional bins were used for the intervals $X_1<-6$ and $X_1>6$. We have generated a data-set of 20000 data-points and tested the method. The results can be seen in Figure \ref{fig:example2}. It can be seen that the method presented a significant benefit over the two other approaches. As mentioned before, for sufficiently large $|X_1|$ the threshold is $-\infty$. This is since $\sigma_i^+>\sigma_i^-$ and the benefit-cost ratio admits a minimum. When the desired benefit-cost ratio is below the minimum possible value it is always desirable to trade more TPR for more FPR. Notice that in those bins the calculation of h(x) is useless and can be avoided, therefore reducing computation resources needed.   
 
 Direct comparison to AUC optimization methods (like \cite{herschtal2004optimising}) is inappropriate. This is since it is highly sensitive to the hypothesis set for which the AUC is optimized.  It is clear from Figure \ref{fig:threshold-ex2} that in spite the fact that our base classifier is linear, no linear classifier can achieve decent performance. Optimizing the AUC over a different hypothesis class may produce better results than ours. However, using this classifier as our base classifier and employing OER may improve it even further or at least will not reduce its performance.
 
 
\subsection{Similar effect on the score variance} \label{Constant Variance}

In some cases, we can assume that the auxiliary features affect only the expectation of the score and do not affect the variance of positive and negative samples. Formally, $\forall i , \sigma_i^-=\sigma_i^+$ .
 In this case  problem (\ref{gen_prob}) can be solved directly. 
 The solution to problem (\ref{gen_prob}) is given by:
 \begin{equation} \label{easy_solution}
 k_i=\frac{{\sigma_i^+}^2[\log\frac{p_i^-}{p_i^+}+\lambda]}{\mu_i^+-\mu_i^-}+\frac{\mu_i^++\mu_i^-}{2},
 \end{equation}
 where $-\infty<\lambda<\infty$
 
   The ROC curve can then be derived by calculating the optimal threshold for different values of $\lambda$ ranging from $-\infty$ to $\infty$. 
 %
  
 \begin{remark}
 For this special case the extension to an infinite number of bins is straightforward. Instead of fitting a Gaussian model to each bin it is possible to estimate some functions $\mu^+(\tilde{x})$ and $\mu^-(\tilde{x})$ that represent the mean score as a function of the features for the positive and negative examples, respectively. Similarly,  the functions $\sigma^+(\tilde{x})$,$\sigma^-(\tilde{x})$,$p^+(\tilde{x})$ and $p^-(\tilde{x})$ should be estimated. All of these functions can be estimated using conventional parametric estimation  methods (For example, maximizing the log likelihood). The optimal threshold for each example can then be calculated using (\ref{easy_solution}) by substituting $\mu^+_i$ by $\mu^+(\tilde{x})$, $\mu^-_i$ by $\mu^-(\tilde{x})$ and so on. 
 \end{remark}
 
 \paragraph{Example 1 revisited}

\begin{figure}
   \centering
   \centering
   \captionsetup{justification=centering,margin=1cm}
   \includegraphics[angle=0,
      width=\linewidth,height=3in]{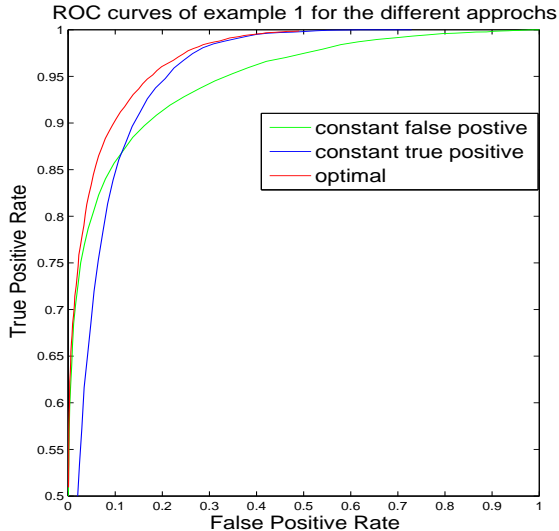}
      \caption{ROC curve of Example 1 }
      \label{fig:example1}
   \end{figure}

 We  used the described method on Example 1. We divided $X_1$ values into 8 bins, where $x\in A_i$ if $0.5+0.5i<X_1<1+0.5i$. It follows that $\mu_i=0.75+0.5i$. Notice that we neglected the fact that $\sigma_i^+ \neq \sigma_i^-$. We have generated a data-set of 20000 data-points and tested the method. The results are presented  in Figure \ref{fig:example1}. It can be seen that the method presented a significant benefit over the two other approaches. Notice also that the derived ROC curve outperforms the convex hull of the two other methods, therefore outperform the ROCCH method.
 
 \section{Finding good features to apply error redistribution on} \label{sec:features}
 
 One important question that arises in the context of OER is how to choose auxiliary features that  provide the most benefit.  Using features that do not contain relevant information may degrade performance due to over-fitting.  
 The simplest approach is probably to use knowledge about the domain of the problem and consider features that may impact the problem difficulty. In image classification this can be for example picture's size, lightning conditions etc. In speaker verification difficulty is often related to the type of recording device. As the quality of the recording gets better it is easier to classify. Using the type of recording device as an auxiliary feature seems natural for this setting.  Other examples include document length in spam filtering, channel characteristics in communication, distance from target in remote sensing and many more.
 
  Another obvious approach is to enumerate over potential options. For each feature apply OER, then calculate the derived ROC and choose the features that gives the most benefit. Sufficient estimation of the ROC however requires large amount of labelled data. In certain cases, labelled data are scarce and therefore the estimation of the ROC is prone to errors. 
 
  An alternative approach is to use the modelling process to uncover potential auxiliary features. Looking at the benefit-cost ratio provides us with the necessary insight about elements of the model that impact performance.     
 One measure that can be proposed  is the difference in separation difficulty. The separation difficulty (SD) is defined by the number of standard deviations between the mean of positive and negative examples. Namely the quantity
 \begin{equation*}
 SD_i=(\mu_i^+-\mu_i^-)/(\sigma_i^+\sigma_i^-).
 \end{equation*}
The difference in separation difficulty can then be defined as $var(SD_i)$.  The variance is taken with respect to the data's distribution. A large difference causes significant bending of the curve for different operating points. While this does not guarantee  significant benefit on the ROC it implies a potential for such benefit. Example 1 demonstrates the feasibility of this measure. 
 
 Another measure is the difference in the prior. The prior of bin $i$ ( denoted by $\mathcal{P}_i$) can be defined as
 \begin{equation*}
 \mathcal{P}_i=\log(p_i^+ \sigma_i^-/ (p_i^-\sigma_i^+)).
 \end{equation*}
 As before, the difference in prior can be taken to be $var(\mathcal{P}_i)$ where the variance is taken with respect to the data distribution.  A large difference indicates that there might be a potential for significant benefits. Example 2 demonstrates the feasibility of this measure.
 
 
 Those measures allow to establish a feature selection mechanism. First, enumerate over possible features, for each feature, partition the space into bins  and measure the difference in separation difficulty and difference in prior. Only features for which those measures exceed some threshold should be used for OER. In the spirit of supervised PCA \cite{bair2006prediction}, further reduction in the feature space's dimension can be achieved by using only the few main  principal components of the remaining features.   The resulting feature space can then be divided into bins and OER can be applied.  
 
 It is important to calibrate the number of bins to the amount of training data available. Using too few bins leads to a mismatch between the data and model and therefore sub-optimal performance (which may even be worse than the original classifier). Using too many bins may lead to over-fitting. We advise using cross-validation in order to optimize the number of bins. 
 
\section{Simulation Results} \label{Simulation Results}
\begin{figure}
    \centering
    \centering
    \captionsetup{justification=centering,margin=1cm}
    \includegraphics[angle=0,
      width=\linewidth,height=2in]{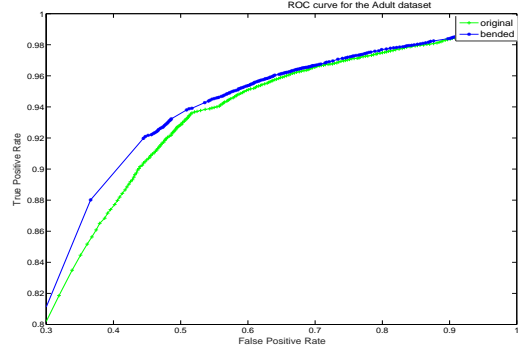}
      \caption{ROC curve for the Adult dataset, $20.33\%$ improvement in AUC }    
      \label{fig:adult_bending}
   \end{figure}

\begin{figure}
    
     \centering
     \captionsetup{justification=centering,margin=1cm}
  \includegraphics[angle=0,
    width=\linewidth,height=2in]{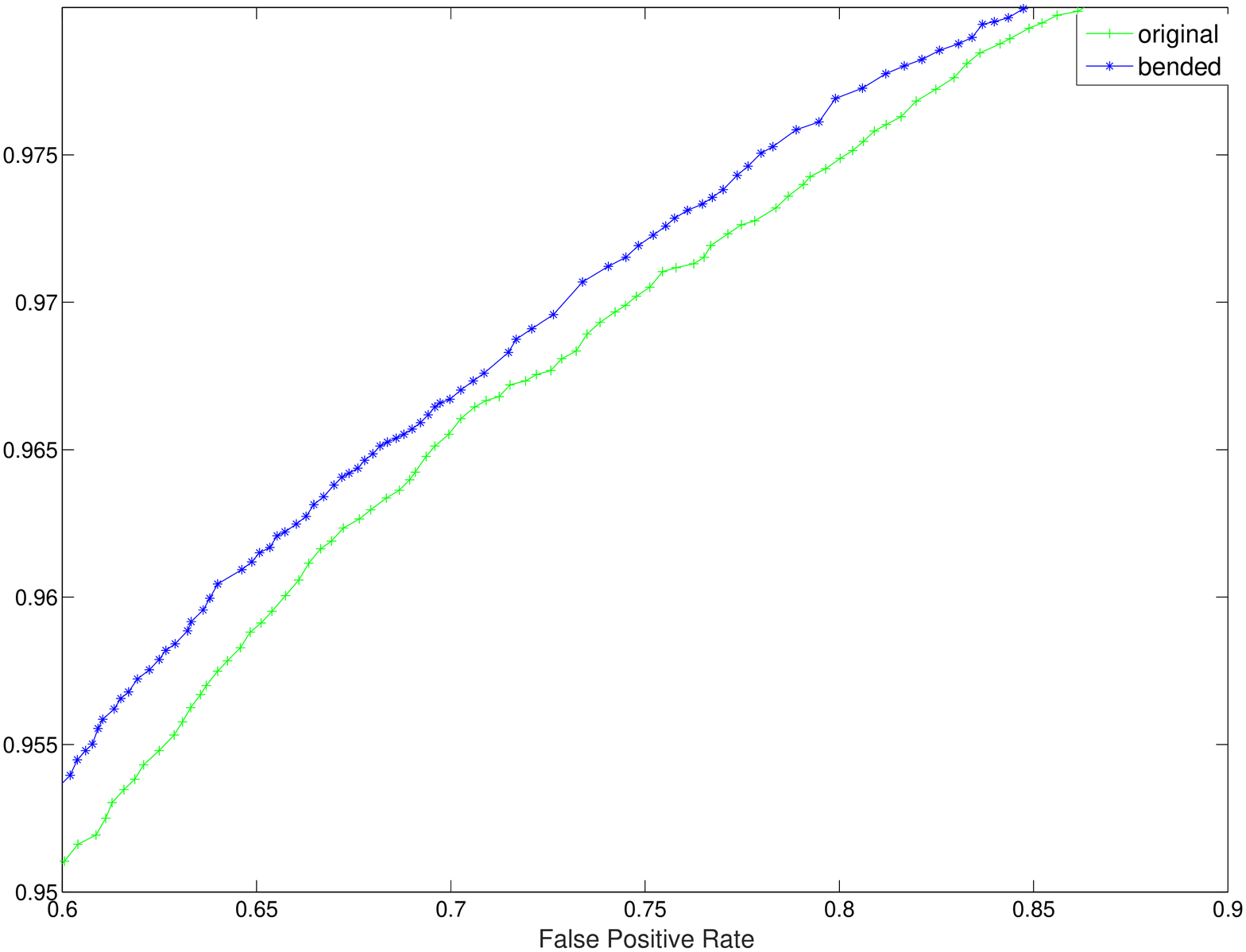}
    \caption{Zoom in on the ROC curve for the Adult dataset}
    \label{fig:adult_bending-zoom}   
    \end{figure}

In addition to the results described earlier on synthetic examples we demonstrate our method's potential benefit on real-life data .
First we tested the method  using the UCI ``Adult'' dataset \cite{Lichman:2013}. 
In this dataset the goal is to predict whether a person's income exceeds $50K/yr$ based on census data.
 We have used SVM as the base classifier. As ab auxiliary feature we selected number of years of education. This selection was made by reviewing the difference in separation difficulty and the difference in prior of all available features as explained in section 4. It is possible that choosing more then one auxiliary feature will improve the results.  Figure \ref{fig:adult_bending}  shows the  derived ROC curves. Figure \ref{fig:adult_bending-zoom} shows a zoom-in of the ROC. It is clearly visible that the derived ROC curve is always better then the original. The AUC improves from 0.878 for the baseline SVM to 0.9028 for the derived classifier, $20.33\%$ improvement. It should be noted that since some of the input features are categorical the ROC curve is highly sensitive. The results shown are averaged over ten-fold cross-validation. Note that on all conducted experiments OER outperforms the original classifier (0.001 p-value with the sign test).
 
\begin{figure}
    \centering
  \includegraphics[angle=0,
    width=0.8\linewidth,height=2in]{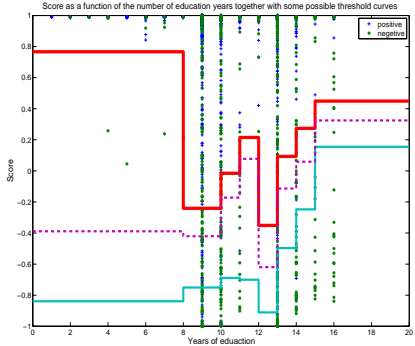}
    \caption{Score as a function of the number of education years together with some possible threshold curves }
    \label{fig:adult_threshold}
    \end{figure}

 Taking a closer look at the data distribution and the derived thresholds shows that the improvement is made by keeping the threshold in the ``easy'' bins low and increasing it on the more ``difficult'' bins. It can be seen that the benefit arise even though the data distribution isn't Gaussian. It is possible that using a different distribution for modelling will produce better results. The data distribution as well as three possible thresholds can be seen on figure \ref{fig:adult_threshold}.
 

 Second, we used OER for the task of object recognition. The task at hand is finding a certain object (person, car, dog, etc.) inside a picture. For that purpose, multiple bounding boxes (BB) are extracted from the picture. A classifier assign a score for each of the BB. Detection is made using some threshold on this score. For simplicity we have tested only the ``classification" stage of this problem. 

From the PASCAL (\cite{Everingham10}) data-base, positive examples of several classes of objects were extracted (only the bounding box which contains the object). From the same data-base, 100000 background examples were taken (from 10 different pictures). Each example was scored using the state of the art Discriminatingly Trained Deformable Part Model classifier   \cite{voc-release5,lsvm-pami}. This classifier models the object as composed out of a set of parts (for example a person is composed out of  head, hands, body, etc.). The classifier then matches the content of the bounding box with all possible orientations of the modelled object and its parts. It is known that the size of the bounding box affects the performance of this classifier significantly \cite{hoiem2012diagnosing}. 

\begin{figure}
  \centering
  \includegraphics[angle=0,
  width=\linewidth,height=3in]{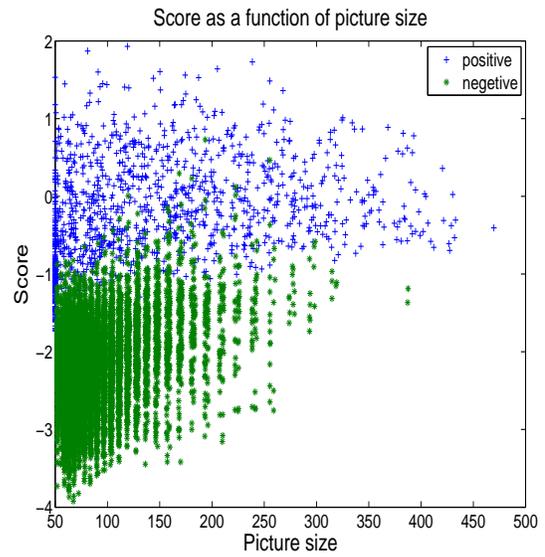}
  \caption{Score as a function of the bounding box size }
  \label{fig:score}
  \end{figure}

The size of the bounding box was used to divide the data into 4 bins.  For each bin the expectation and standard deviation of the positive and negative examples were estimated as well as $p^+$ and $p^-$.  The scores for the class ``person'' as a function of size can be seen in Figure \ref{fig:score}.

 Two effects are notable. First, the bigger the BB (higher resolution) the higher the score.  It can be seen that the effects on positive and negative examples are roughly the same in expectation but for larger BB the variance of the positives decrease while the variance of the negatives remain roughly the same. Second, Since the data-base is constructed from partitioning of pictures, it contains a high number of small BBs and a low number of large BBs. The positive examples however are distributed roughly uniform over size. This causes the change of prior to be rather large. 

\begin{figure}
  \centering
  \centering
  \captionsetup{justification=centering,margin=1cm}
  \includegraphics[angle=0,
    width=\linewidth,height=3in]{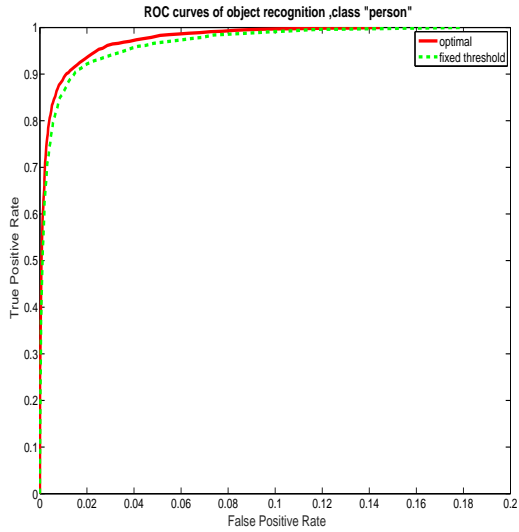}
    \caption{ROC curve of object recognition, class ``person'' }    
    \label{fig:object}
    \end{figure}

\begin{figure}
    \centering
   \captionsetup{justification=centering,margin=1cm}
\includegraphics[angle=0,
  width=\linewidth,height=3in]{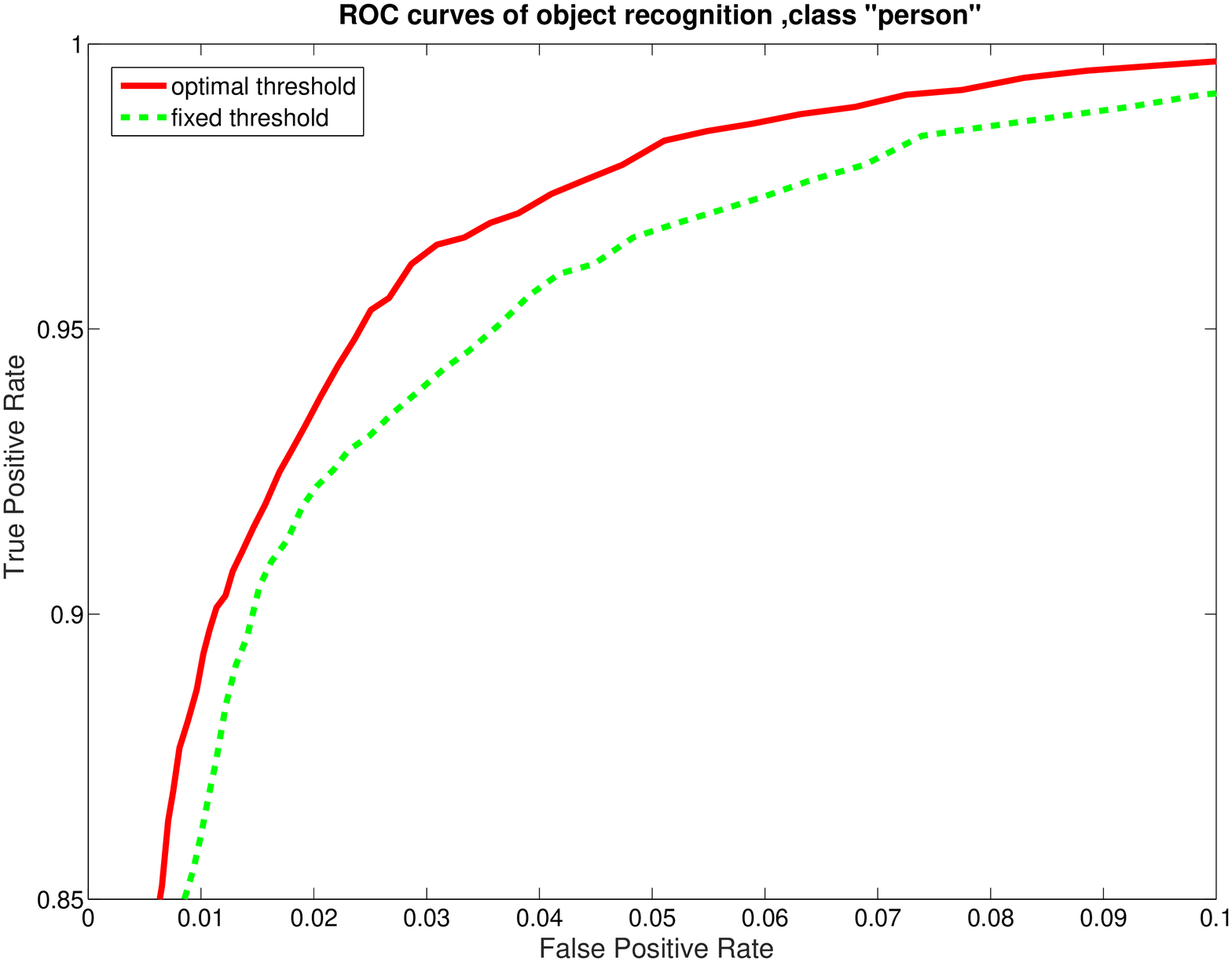}
  \caption{Zoom in on the ROC curve of object recognition, class ``person'' }
  \label{fig:object-zoom}   
  \end{figure}

Optimal thresholds were calculated using OER and the results were compared to using a fixed threshold. The area under the curve (AUC) was used as a performance measure.
 The results are summarised in Table \ref{table:object}. As can be seen, substantial benefit (around 20\% improvement) arises from using OER. Further examination of the benefit shows that for a very low FPR modelling errors start to affect and benefit is minor. For a very high FPR there is not much room for improvement. In between, there is a substantial area in which benefit arise. Figures \ref{fig:object}  shows the  derived ROC curves for the class ``person''. Figure \ref{fig:object-zoom} shows a zoom-in of the ROC of the area in which the benefit is maximal. This improvement is done using only the picture size as a feature. This feature boost the base classifier's performance although it holds little to non discriminative information. Ten-fold cross-validation was performed. 
 Recently the validity of AUC for model comparison was questioned \cite{hand2009measuring}. While for simplicity we do use AUC as a performance measure our method improve the entire ROC curve. Since the improved ROC curve dominates the original one, other measures will also likely to show improvement. 
 
 Note that on all conducted experiments OER outperforms the original classifier (0.001 p-value with the sign test).  Notice also that we have used only a few bins (four) and a simplistic modelling as Gaussian. We believe that by using more complex features and more complex models this results can be even further improved. 


\begin{table}
\caption{Simulation results for several object classes }
\begin{center}
\begin{small}
\begin{tabular}{|p{1.5cm}|c|c|c|c|}
\hline
class & person &dog  &car & chair  \\
\hline
Number of positive examples&2358&253&625&400\\
\hline
Fixed thresholds AUC&0.98663&0.97827&0.99292&0.99540\\
\hline
OER AUC &0.99043&0.98329&0.99411&0.99648\\
\hline
Improvement in $1-AUC$&\bf{28.42\%}&\bf{23.1\%}&\bf{16.81\%}&\bf{23.48\%}\\
\hline
\end{tabular}
\end{small}
\end{center}
  \label{table:object}
\end{table}

\section{Conclusion} \label{section:conclusion}
In this work we  present a novel approach for improving the ROC curve of existing classifiers. We believe that this method should become a standard tool in ROC analysis and can enhance essentially any classifier. The method presented is general and may provide substantial benefit for any application: as long as there is sufficient data to mitigate overfitting, anyone who considers ROC optimization should try to ``bend the curve" since there is nothing much to lose from it, and potentially much to gain.

We suggest three natural directions for further research:
First, the method presented takes a two step approach. Start with modelling the data and then find optimal threshold curve according to this model. The model is used to derive the benefit-cost ratio. An alternative approach is to use empirical estimates of the benefit-cost ratio directly. The effect of such an approach is unclear. On the one hand, it may improve performance whenever a parametric model is in-adequate to describe the data. On the other hand, it may increase over-fitting. 

Second, accurate estimation of the model's parameters requires a large amount of labelled data. This is especially true when the number of prospective features is large. Partitioning the space into too many bins may lead to a faulty model. An interesting open question is how to optimally partition the feature space. 

 Third, in some scenarios it may be preferable to use a different optimization problems than (\ref{gen_prob}). For example, in multi-view problems several classifiers, each with a different feature-space, are fused into a single classification output. It may be interesting to jointly optimize the threshold curves of those classifiers. 

\bibliographystyle{plain}
\bibliography{mybib2}{}
\end{document}